# NileTMRG at SemEval-2017 Task 4: Arabic Sentiment Analysis

**Samhaa R. El-Beltagy**[1], **Mona El Kalamawy**[2], **Abu Bakr Soliman**[1]
[1]Center for Informatics Sciences
Nile University, Juhayna Square, Sheikh Zayed City, Giza, Egypt
[2]Faculty of Computers and Information,
Cairo University, Ahmed Zewail St, Giza, Egypt
samhaa@computer.org, mona.elkalamawy@fci-cu.edu.eg, ab.soliman@nu.edu.eg

## Abstract

This paper describes two systems that were used by the NileTMRG for addressing Arabic Sentiment Analysis as part of SemEval-2017, task 4. NileTMRG participated in three Arabic related subtasks which are: Subtask A (Message Polarity Classification), Subtask B (Topic-Based Message Polarity classification) and Subtask D (Tweet quantification). For subtask A, we made use of our previously developed sentiment analyzer which we augmented with a scored lexicon. For subtasks B and D, we used an ensemble of three different classifiers. The first classifier was a convolutional neural network for which we trained (word2vec) word embeddings. The second classifier consisted of a MultiLayer Perceptron while the third classifier was a Logistic regression model that takes the same input as the second classifier. Voting between the three classifiers was used to determine the final outcome. The output from task B, was quantified to produce the results for task D. In all three Arabic related tasks in which NileTMRG participated, the team ranked at number one.

## 1 Introduction

Because of the potential impact of understanding how people react to certain products, events, people, etc., sentiment analysis is an area that has attracted much attention over the past number of years. The consistent increase in Arabic social media content since 2011 (Neal 2013)(Anon 2012)(Farid 2013) resulted in increased interest in Arabic sentiment analysis. Lack of Arabic resources (datasets and lexicons), initially hindered research efforts in the area, but the area gradually gained attention, with research effort either focusing on building missing resources (El-Beltagy 2016; Refaee & Rieser 2014; El-Beltagy 2017), or on experimenting with different classifiers and features while creating needed resources as is briefly described in the related work section.

In this paper we present our approach to addressing the following three SemEval related sentiment analysis subtasks (Arabic):

A) Message Polarity Classification: given a tweet/some text the task is to determine whether the tweet reflects positive, negative, or neutral sentiment.
B) Topic-Based Message Polarity Classification: given some text and a topic, determine whether the sentiment embodied by the text is positive or negative towards the given topic.
D) Tweet quantification: given a set of tweets about a given topic, estimate their distribution across the positive and negative classes.

Two systems have been used to address these tasks. The first system is a slightly altered version of that presented in (El-Beltagy et al. 2016). The second is composed on an ensemble of three different classifiers: a convolutional neural network(Kim 2014), a Multi-Layer Perceptron, and a Logistic regression classifier.

The rest of this paper is organized as follows: section 2 presents a brief overview of related work, section 3 describes the datasets used for training, section 4 overviews the developed systems, while section 5 presents the evaluation results, and section 6 concludes the paper.

# 2 Related Work

## 2.1 Task A

Research in Arabic Sentiment analysis has been gaining momentum over the past couple of years. The work of (El-Beltagy & Ali 2013) outlined challenges faced for carrying out Arabic sentiment analysis and presented a simple lexicon based approach for the task. (Abdulla et al. 2013) compared machine learning and lexicon based techniques for Arabic sentiment analysis on tweets written in the Jordanian dialect.

The best obtained results were reported to be those of SVM and Naive Bayes. The work presented in (Shoukry & Rafea 2012) targeted tweets written in the Egyptian dialect and was focused on examining the effect of different preprocessing steps on the task of sentiment analysis. The authors used a SVM classifier in all their experiments. (Salamah & Elkhlifi 2014) developed a system for extracting sentiment from the Kuwaiti-Dialect. They experimented with a manually annotated dataset comprised of 340,000 tweets, using SVM, J48, ADTREE, and Random Tree classifiers. The best result was obtained using SVM. (Duwairi et al. 2014) presented a sentiment analysis tool for Jordanian Arabic tweets. The authors experimented with Naïve Bayes (NB), SVM and KNN classifiers. The NB classifier performed best in their experiments. (Shoukry & Rafea 2015) presented an approach that combines sentiment scores obtained using a lexicon with a machine learning approach applied it on Egyptian tweets. The experiments conducted by the authors, show that adding the semantic orientation feature does in fact improve the result of the sentiment analysis task. (Khalil et al. 2015) experimented with various datasets, classifiers and features representations to determine which configurations work best for Arabic sentiment analysis. They concluded that Multi-nominal Naïve Bayes and Complement Naïve Bayes tend to work best, especially when term features are represented using their idf weights.

## 2.2 Tasks B and D

To the knowledge of the authors, no previous work on "Topic-Based Message Polarity Classification" has been attempted on Arabic. The same task has been introduced last year at SemEval-2016, so a number of systems have been developed to address this task in English. For this task, most participants preferred to use a deep learning approach. The Finki team for example (Stojanovski et al. 2016) developed a system composed of a merged convolutional neural network with a gated recurrent neural network. In their system, pre-trained Glove word embeddings were used to represent tweet tokens. The "thecerealkiller" team (Yadav 2016) on the other hand, used only a recurrent neural network. In their system, tweets were minimally preprocessed before being fed to the network. The "LyS" team (Vilares et al. 2016) used a convolutional neural network with support vector machines. They trained the SVM using hidden CNN activations with additional linguistic information. The "Tweester" team (Palogiannidi et al. 2016) used multiple independent classifiers: neural networks, semantic-affective models and topic modeling. Each classifier predicts the result and late fusion is done to generate the final result.

# 3 Used Data

## 3.1 Task A

The organizers provided a total of 3355 sentiment labeled tweets for training and an additional 671 labeled tweets for validation/development. Close examination of the provided labeled tweets revealed that some tweets in the training set were duplicated, sometimes even with conflicting labels. To use those for training, the data was cleaned as follows:

1) All tweets that had conflicting sentiment labels were removed.
2) A single copy of tweets which were duplicated, but had a consistent label, was kept.
3) Any tweet in the training dataset which was found in the development dataset, was deleted.

This brought down the number of training tweets to 2499. Initial experimentation using only this data for training resulted in rather low performance. Additional data in the form of the NBI dataset described in (El-Beltagy et al. 2016) was augmented to this training data after balancing it with some data from the Nile University (NU) dataset also described in (El-Beltagy et al. 2016). In addition, lexicon terms that had a score higher than a 0.8 threshold as determined by the work described in (El-Beltagy 2017) were added as entries to the training dataset with their polarity as a label. The final training dataset consisted of 13,292 tweets/entries.

### 3.2 Tasks B and D

The organizers provided 1322 tweets with topics and their sentiment labels for training and a further 332 tweets for validation/development. The only additional data that we have used, consisted of 4 million Arabic tweets which were used to train a word2vec model (Mikolov, Sutskever, et al. 2013; Mikolov, Corrado, et al. 2013) for use with a convolutional neural network (Kim 2014).

## 4 System Descriptions

### 4.1 The System used for Task A

As mentioned earlier, the system used for task A, is a modified version of the NU sentiment analyzer described in (El-Beltagy et al. 2016). The analyzer was built using a Complement Naïve Bayes classifier (Rennie et al. 2003) with a smoothing parameter of 1 and trained using the 13,292 Arabic tweets described in section 3.1. Complement Naïve Bayes was selected as a classifier based on the work presented in (Khalil et al. 2015). Input to the model consists of feature vector representations of input tweets. Each vector represents unigrams and bigram terms with their idf weights and has an additional set of lexical features for the input tweet that can be summarized as follows:

- *startsWithLink*: a feature which is set to 1 if the input text starts with a link and to a 0 otherwise.
- *endsWithLink*: a feature which is set to 1 if the input text ends with a link and to 0 otherwise.
- *numOfPos*: a count of terms within the input text that have matched with positive entries in our sentiment lexicon.
- *numOfNeg*: a count of terms within the input text that have matched with negative entries in the sentiment lexicon.
- *length*: a feature that can take on one of three values {0,1,2} depending on the length of the input text. The numbers correspond to very short, short and normal.
- segments: a count for the number of distinct segments within the input text.
- *endsWithPostive*: a flag that indicates whether the last encountered sentiment word was a positive one or not.
- *endsWithNegative*: a flag that indicates whether the last encountered sentiment word was a negative one or not.
- *startsWithHashTag*: a flag that indicates whether the tweet starts with a hashtag.
- *numOfNegEmo*: the number of negative emoticons that have appeared in the tweet.
- *numOfPosEmo*: the number of positive emoticons that have appeared in the tweet.
- *endsWithQuestionMark*: a flag that indicates whether the tweets ends with a question mark or not.

In addition to these features that were originally described in (El-Beltagy et al. 2016), we have made use of our newly created weighted sentiment lexicon (El-Beltagy 2017) to add two additional features:

- *negScore*: a real number that represents that score of all negative terms in the input text.
- *posScore*: a real number that represents that score of all negative terms in the input text

It is worth noting that these two features replaced the *negPercentage* and *posPercentage* features described in (El-Beltagy et al. 2016). As stated in (El-Beltagy 2017), an amplification factor for these scores does enhance the classifier's performance. When experimenting using the supplied validation dataset, using these additional two features versus not using them, did make a difference. Additionally, the validation dataset was used to:

- Determine whether the removal of any of the listed features would impact system performance positively. In the end, all features were kept.
- Experiment with various training dataset combinations.
- Determine the amplification factor to use.

In task A, we only used the validation dataset for fine tuning. The model that performed best on the validation dataset, was the model used to generate sentiment labels for the test data. Additional pre-processing steps included character normalization, mention normalization, elongation removal, emoticon replacement, and light stemming. Further details on each of these steps can be found in (El-Beltagy et al. 2016).

### 4.2 The System used for Tasks B and D

The approach followed for addressing task B, was one in which three independent classifiers were built using the provided training data. Voting among the three classifiers, determined the final label for any given tweet. Task D is in fact highly dependent on task B, so we cannot say that we built a separate system for that task. Instead, the output of task B is simply counted and quantified to produce the output of task D. While validation data was used to fine tune this system, it was also augmented to the training data to build the final model which was used to generate labels for the test data. In the following subsections we described each of the used classifiers.

#### 4.2.1 The convolutional Neural Network

The CNN model (Kim 2014) that we employed, is composed of 2 convolutional layers with filters of window sizes 3 and 4. A max pooling layer was added to extract the most important features, then a fully connected layer of size 25 nodes was connected to the pooling layer. A softmax layer generates the predicted class label. We have implemented this model using Keras (Chollet 2015).

**Tweet Preprocessing**

The following preprocessing steps were carried out on tweets that were used with this classifier:

- Hyperlinks and diacritics were removed
- Elongation was eliminated (ex. Yessss -> Yes)
- Text was normalized: (آ ,إ ,أ) → (ا) , (ي)→(ى) , (ة) → (ه)
- Positive emoticons were replaced by the word “حب” (love), and negative emoticons by “غضب” (anger).
- The word “حب” (love) was added beside words that indicate positivity like: (شجاع, عظيم, ممتاز) (brave, excellent, great) and the word “غضب” (anger) next to words that indicate negativity like (حطم, يسئ, هابط) (broke/destroyed, harms, lowly)
- If the target topic was found in the tweet’s text, it was replaced with a static keyword

**Input to the CNN**

Each word in a tweet is represented by its embeddings vector, the dimensionality of which is 100. The whole tweet is thus represented by the aggregate of vectors for each of its words which is essentially a matrix. As a CNN requires a fixed size input, we chose to set the matrix size to the max tweet length (which was 35 words in our case). For tweets smaller than the max, word embeddings were centered in the middle and padded with zeros around. Note that the number of rows is constant (embedding vector length). If a word wasn’t found in the word2vec model, we set its embeddings vector to random numbers.

#### 4.2.2 The Multi-Layer Perceptron & Logistic Regression Classifiers

We have grouped these two classifiers together as they essentially take in the same input features. Tweets are preprocessed for these classifiers in exactly the same way they are preprocessed for the CNN except that we don’t replace the target topic with a static keyword.

**Input to the Classifiers**

The input to both the multi-layer perceptron and logistic regression classifiers is a set of feature vectors representing input tweets. A feature vector is composed to a bag of words representation of the tweet in addition to the following features:

1) **The overall sentiment** of the tweet (regardless of the target). The NU Sentiment analyzer (El-Beltagy et al. 2016) was used to set this feature. Since two teams were working in parallel on tasks A and B respectively, the team working on task B simply used the older version of the system with old training data. We believe that using the modified version that was described for task A, might yield better results.
2) **The number** of positive/negative words found in the tweet.
3) A flag indicating the presence of positive emoticons
4) A flag indicating the presence of negative emoticons
5) The position of the target in the tweet
6) A flag to indicate if there are positive **terms** around the target topic
7) A flag to indicate if there are negative **terms** around the target topic
8) A flag to indicate if there positive/negative **words** around the target?

9) The number of positive terms in the first half of the tweet
10) The number of negative terms in the first half of the tweet
11) The number of positive term in the second half of the tweet
12) The number of negative terms in the second half of the tweet

Features 1, 2, 6, and 7 were amplified to emphasize their importance.

# 5 Results

## 5.1 Task A

The supplied test data for task A consisted of 6100 unlabeled tweets. These tweets were classified using the system described in section 3.4. Based on the results supplied by the organizers of the task (Rosenthal et al. 2017), our system ranked at number 1 as shown in table 1.

| # | System | *P* | $F_1^{PN}$ | *Acc* |
|---|---|---|---|---|
| 1 | **NileTMRG** | $0.583_1$ | $0.610_1$ | $0.581_1$ |
| 2 | SiTAKA | $0.550_2$ | $0.571_2$ | $0.563_2$ |
| 3 | ELiRF-UPV | $0.478_3$ | $0.467_4$ | $0.508_3$ |
| 4 | INGEOTEC | $0.477_4$ | $0.455_5$ | $0.499_4$ |
| 5 | OMAM | $0.438_5$ | $0.422_6$ | $0.430_8$ |
| 6 | LSIS | $0.438_5$ | $0.469_3$ | $0.445_6$ |
| 7 | Iw-StAR | $0.431_7$ | $0.416_7$ | $0.454_5$ |
| 8 | HLP@UPENN | $0.415_8$ | $0.320_8$ | $0.443_7$ |

Table 1: Results for Subtask A "Message Polarity Classification", Arabic. The systems are ordered by average recall ρ

## 5.2 Task B

The supplied test data for task B consisted of 2757 tweets with each having a given topic. Voting amongst the 3 classifiers described in section 4.2, determined the final label for the topic (positive or negative). The results for our system, which ranks us at number 1, are shown in table 2.

| # | System | *P* | $F_1^{PN}$ | *Acc* |
|---|---|---|---|---|
| 1 | **NileTMRG** | $0.768_1$ | $0.767_1$ | $0.770_1$ |
| 2 | ELiRF-UPV | $0.721_2$ | $0.724_2$ | $0.734_2$ |
| 3 | ASA | $0.693_3$ | $0.670_4$ | $0.672_4$ |
| 4 | OMAM | $0.687_4$ | $0.678_3$ | $0.679_3$ |

Table 2: Results for Subtask B "Tweet classification according to a two-point scale", Arabic. The systems are ordered by average recall ρ

To better understand the performance of the voting system, the performance of the individual classifiers which participated in the voting process, is presented in table 3.

| # | System | *ρ* | $F_1^{PN}$ | *Acc* |
|---|---|---|---|---|
| 1 | MLP | 0.728 | 0.720 | 0.720 |
| 2 | **LR** | 0.762 | 0.759 | 0.761 |
| 3 | CNN | 0.747 | 0.737 | 0.738 |

Table 3: Individual results of each of the used classifiers

## 5.3 Task D

Test data for task D was identical to that for task B. In order to obtain results for task B, a script was created that takes in the input from task B and converts it to the required output for task D. The results for our system, which ranks us at number 1, are shown in table 4.

| # | System | *KLD* | *AE* | *RAE* |
|---|---|---|---|---|
| 1 | **NileTMRG** | $0.127_1$ | $0.170_1$ | $2.462_1$ |
| 2 | OMAM | $0.202_2$ | $0.238_2$ | $4.835_2$ |
| 3 | ELiRF-UPV | $1.183_3$ | $0.537_3$ | $11.434_3$ |

Table 4**:** Results for Subtask D "Tweet quantification according to a two-point scale", Arabic. The systems are ordered by their KLD score (lower is better)

# 6 Conclusion

In this paper, we have briefly described our submissions to SemEval task-4, subtasks A, B, and D (Arabic). For task A, we have augmented our previously developed sentiment analyzer with a scored lexicon and trained it using a little more that thirteen thousand tweets. For tasks B and D, we have used three independent classifiers and employed a voting strategy to determine the final label for a given topic. The evaluation results for all three tasks have ranked our systems at number one.